# Advancing Vision-based Human Action Recognition: Exploring Vision-Language CLIP Model for Generalisation in Domain-Independent Tasks

Utkarsh Shandilya[1], Marsha Mariya Kappan[2], Sanyam Jain[3†], and Vijeta Sharma[4]

[1] Department of Computer Science and Engineering, Central University of Haryana, Jant-Pali, Haryana, India
utyk1001169@gmail.com
[2] University of New South Wales, Sydney NSW 2033, Australia
m.marsha_mariya_kappan@unsw.edu.au
[3] Dept. of Computer Science and Communication, Østfold University College, 1783 Halden, Norway
sanyamjaincs@gmail.com
[4] Norwegian University of Science and Technology, 2815 Gjøvik, Norway
vijeta.sharma@ntnu.no

**Abstract.** Human action recognition is pivotal in healthcare and medicine, enabling applications such as patient behaviour analysis, fall detection, surgical robot monitoring, and procedural skill training. While traditional models like CNNs and RNNs have demonstrated success, their capacity for recognising diverse and complex actions remains limited. Recent advances in vision-language models, particularly the transformer-based CLIP model, show promise for generalising predictions from video data. In this paper, we evaluate the CLIP model on the UCF-101 action recognition dataset and analyse its performance under three masking strategies: (a) random percentage-based and shape-based black masking at 10%, 30%, and 50%, (b) feature-specific masking to remove bias-inducing features, and (c) isolation masking, where class-specific features are selectively preserved and labelled. Our findings reveal inconsistent behaviour and frequent misclassifications, particularly when key visual patterns are disrupted. To address these limitations, we introduce class-specific noise, learned through a custom loss function, which enhances the model's focus on class-defining features. This approach improves classification accuracy and confidence, offering a robust, unbiased, precise action recognition solution. We further discuss challenges and propose future directions to extend this approach's generalizability in medical contexts, as this generalisation works as domain-independent scenarios. The code is available at https://github.com/s4nyam/HAR-CLIP

## 1 Introduction

Vision-Language Models (VLMs) have emerged as a transformative approach in the field of computer vision, leveraging the rich relationships between visual data and textual descriptions. Unlike traditional vision-only models that focus solely on image data, VLMs integrate both visual and textual information, enabling more comprehensive understanding and interpretation of the content [35, 5]. Prominent examples of VLMs include CLIP [20] and ALIGN [14], which utilize large-scale image-text pairs for pre-training. These models have shown remarkable success in various tasks, such as image classification [1], zero-shot learning [23], and visual question answering [3], by capturing correspondences between images and their textual descriptions.

In the realm of action recognition, VLMs have been employed to improve the understanding of dynamic human behaviors within videos [30, 11, 34, 2, 29]. Action recognition tasks aim to classify human actions in video sequences, posing significant challenges due to the variability of actions, background noise, and occlusions. Despite the advancements brought by VLMs, models still encounter issues related to generalization [21], particularly when class-specific features are improperly generalized or neglected—a phenomenon known as *label dispersion* [4]. This challenge is exacerbated by the complexities inherent in action recognition tasks, where distinguishing between similar actions often requires focusing on subtle, class-defining features [32, 33, 16].

To effectively evaluate the robustness of VLMs in action recognition, it is crucial to explore various masking and perturbation strategies. By applying different masking techniques—such as percentage-based masking or feature-specific masking—we can assess how well models maintain performance under conditions that obscure or alter significant aspects of the input data [6, 12, 24]. These perturbations serve as a valuable criterion for understanding biases in feature learning and the generalization capabilities of VLMs, as they reveal the models' reliance on specific features and their ability to adapt to noisy or incomplete information.

Current noise-based augmentation methods have been explored to address the issues of bias and label dispersion in action recognition [9, 10]. These techniques introduce noise into the input data, effectively simulating variations and perturbations that the model might encounter in real-world scenarios. By training VLMs on augmented data that includes class-specific noise, we can enhance the models' focus on relevant features, ultimately improving their classification accuracy and resilience against misclassifications.

These models are also advancing day by day, making them suitable for complex healthcare and medical applications, such as chest radiographs summarization using Medical Vision-Language Models [26], generating medical captions and addressing complex medical queries [18]. Among the most notable VLMs, the CLIP model developed by OpenAI has demonstrated exceptional generalisation capabilities, enabling it to perform various visual recognition tasks across diverse domains [20]. While CLIP's pre-trained transformer-based architecture excels at image classification and retrieval tasks, its potential for vision tasks [19, 31, 22], such as human action recognition [28], Very little research has been done so far on generalisation in domain independence. CLIP's ability to generalise across visual contexts makes it a promising candidate for tackling action recognition-related challenges [25], particularly when dealing with real-time predictions and diverse healthcare settings [36, 15].

This paper investigates the applicability of CLIP in human action recognition tasks to make it generalise in any domain. Since there is a scarcity of daily living healthcare and medical domain datasets, we used the daily living popular benchmark HAR dataset with the objective of this experiment, which has the potential to scale up to the healthcare future HAR dataset. We focus on CLIP based HAR's robustness against masking strategies and evaluating their generalisation performance through the introduction of a novel class-specific noise augmentation technique. Our findings aim to contribute to the understanding of biases in action recognition tasks and highlight the potential of VLMs in overcoming existing challenges.

## 2   Related Work

In a recent paper [21], explore the limitations of VLMs when it comes to low-level vision tasks that require precise spatial understanding. Their BlindTest benchmark, which includes simple tasks such as determining whether two circles overlap or counting line intersections, demonstrated that state-of-the-art VLMs, including models like GPT-4V and Gemini-1.5, achieved an average accuracy of only 58.57%, far below the expected human-level performance. This finding highlights a significant gap in the ability of VLMs to handle basic visual perception tasks, especially when required to process geometric shapes and spatial relationships.

While VLM based CLIP models excel in zero-shot classification and retrieval tasks, recent research [8], has revealed that their learned representations are less effective for dense prediction tasks such as object detection, semantic segmentation, and depth estimation. In s study by [7], evaluated the effectiveness of CLIP for the task of Medical Visual Question Answering (MedVQA) and presents PubMedCLIP model, a fine-tuned version of CLIP for the medical domain based on PubMed articles. Authors [27], presented that enhancing the quality of captions in image-text datasets significantly improves CLIP's visual representations, leading to better performance on downstream dense prediction tasks. They demonstrated CLIP's efficiency with well-aligned image-text pairs in semantic segmentation and depth estimation, leveraging Masked Image Modeling (MIM) pretraining methods.

The study also underscores the problem of VLMs' reliance on "late fusion" mechanisms, where visual features are extracted prior to task-specific interpretation, often leading to misinterpretation in scenarios involving fine-grained spatial details. This shortcoming becomes particularly relevant in tasks like action recognition, where subtle, class-specific features are essential for accurate classification. This [21] work serves as a foundation for understanding the gaps in current VLMs, especially regarding their ability to maintain focus on critical visual patterns. Building on this insight, our work further explores the impact of different masking strategies on VLM performance and proposes class-specific noise augmentation as a solution to mitigate label dispersion and improve robustness in action recognition tasks.

## 3   Background

This section overviews the terminologies and definitions used in our study in the context of the CLIP model for Human Action Recognition.

### 3.1   CLIP Model

The CLIP (Contrastive Language-Image Pretraining) model, developed by OpenAI, is a Vision-Language Model (VLM) designed to learn a joint embedding space for images and text. The architecture is built on two main components:

– **Visual Encoder:** This type of encoder is typically based on a transformer or a convolutional neural network (CNN) architecture (e.g., ResNet or ViT). It processes input images into a dense feature representation and extracts visual features that capture semantic and contextual information from the image.
– **Text Encoder:** It is a transformer-based language model (like GPT or similar architectures). It converts input text descriptions into a corresponding feature representation in the same embedding space as the visual encoder.

### 3.2   Contrastive Learning

CLIP is trained using a large dataset of image-text pairs. The model optimizes a contrastive loss function that aligns image embeddings and their corresponding text embeddings in a shared latent space. This enables zero-shot learning, where the model can generalize to unseen tasks by understanding textual descriptions of the task. During training, the model learns to associate visual patterns with corresponding textual descriptions, bridging the gap between vision and language modalities.

### 3.3   Human Action Recognition with Pretrained Generalization

Human action recognition involves identifying and categorizing actions performed in videos. CLIP's architecture and capabilities make it highly applicable to generalize across domains in complex or real-world environments. For example, using zero-shot and few-shot learning, CLIP can classify actions without extensive task-specific training by leveraging its ability to interpret textual descriptions of actions. For instance, textual prompts like "a person performing surgery" or "a patient falling" guide the model in detecting corresponding actions.

### 3.4   Applications in Healthcare

There are several potential applications of the CLIP model in healthcare where our study is highly relevant, such as:
– **Patient Monitoring:** Recognizing activities such as falls, mobility patterns, or therapy exercises.
– **Surgical Skill Assessment:** Identifying precise actions or errors during surgical procedures.
– **Anomaly Detection:** Detecting unusual or critical actions, such as distress signals or unsafe movements.

## 4   Methods

In this work, we implement various methods, each proposed as a task to address the label dispersion problem in action recognition using CLIP. The tasks are defined below along with the corresponding methods.

**Task 1:** Prediction (using CLIP) based comparative analysis of action classes in UCF101 dataset using frequency histogram charts. This method involves extracting frames from the UCF101 video dataset and tasking the model to label each frame according to its respective action class. The frequency histogram is then used to analyze how often the model correctly classifies each action class out of the 101 available categories. The expected result is a high frequency of correct classifications with minimal label dispersion.

**Task 2:** Introduce random-shape and random-percentage perturbation to evaluate VLMs. In this task, random perturbations are applied to the input frames by masking random shapes at various percentages (10%, 30%, and 50%) of the frame's area. The VLMs are then evaluated to observe how often they misclassify actions under these perturbations, using frequency histograms to show the resulting label dispersion. The model's prediction confidence is also measured, with lower confidence and more frequent misclassifications expected as perturbation levels increase.

**Task 3:** Feature-specific masking for each action class. In this task, frames corresponding to each action class are processed through a segmentation model to mask non-class-specific features. For example, in the *cricket shot* action, non-class-specific features like the field, stadium, and boundaries are masked, leaving only the player visible. The model is evaluated in two ways: (1) masking one feature at a time across three randomly chosen non-class-specific features, and (2) masking all three features simultaneously. The frequency histograms are used to measure model performance, showing how the model's predictions shift when specific features are masked.

**Task 4:** Isolation masking. This task builds on Task 3 by masking all non-class-specific features, leaving only the class-defining features visible in each frame. For example, in the *cricket shot* class, only the player making the shot is preserved while all background features are masked. The model is evaluated through frequency histograms to assess how well it performs when only class-specific features are available.

**Task 5:** Class-specific noise augmentation. In this task, class-specific noise is learned as a dictionary for each action class using a custom loss function. This noise aims to help the model focus on retaining key class-defining features. The evaluation will involve measuring the frequency of misclassifications and the confidence levels for each class.

Below are descriptions of key concepts used in the tasks above.

**Frequency Histogram Chart (FHC):** FHCs show the count of labels predicted for each action class in UCF101, helping to visualize label dispersion and misclassifications. The FHCs also display the average confidence of the model for each class, indicating how certain the model is in its predictions. For example, in the *typing* class, if the model frequently misclassifies the action as another label, the FHC will show the count of these misclassifications and the average confidence scores for those predictions.

**Segmentation Model:** In this work, we utilize the pre-trained Segment Anything Model (SAM) [17], developed by Meta AI, for all segmentation tasks. SAM is designed for general-purpose object segmentation and is known for its versatility across diverse datasets and tasks. The model can generate high-quality masks with minimal prompts, making it ideal for segmenting complex frames in the UCF101 dataset. SAM's ability to precisely isolate objects, such as individuals or action-related elements, is crucial for both feature-specific masking and isolation masking tasks. By leveraging SAM's pre-trained capabilities, we ensure that class-specific and non-class-specific features are accurately segmented, facilitating a focused evaluation of the vision-language models' performance in action recognition.

**Class-Specific Noise:** In this work, we introduce class-specific noise as a learnable augmentation technique aimed at enhancing the model's ability to differentiate between action classes by focusing on class-defining features from the UCF101 dataset. The proposed method incorporates triplet loss to ensure that the learned class-specific noise not only highlights key features but also strengthens inter-class separability. The method consists of the following steps:

1. **Class Feature Extraction and Dictionary Creation:** We extract class-specific features from frames of each video in the UCF101 dataset. For each class $c$, a dictionary $\mathcal{D}_c$ is constructed, storing the characteristic features of that class. These features are augmented by learnable noise $\mathbf{N}_c$ for each class, which serves as an additional representation to strengthen the class-distinctiveness.

The feature representation of a class $c$, denoted as $\mathbf{F}_c$, is given by:

$$\mathbf{F}_c = f(\mathbf{X}_c) + \mathbf{N}_c$$

where $f(\mathbf{X}_c)$ represents the extracted features from the model for class $c$, and $\mathbf{N}_c$ is the learnable noise associated with class $c$.

2. **Class-Specific Noise Augmentation:** The noise $\mathbf{N}_c$ is applied to the feature representations during training. This noise, initialized randomly, is optimized during the learning process, becoming a crucial component in the model's understanding of class-specific patterns. The learned noise ensures robustness by perturbing the features, forcing the model to adapt and differentiate classes more effectively. Specifically, the noise-modified feature for a sample from class $c$ can be written as:

$$\mathbf{F}_c^{\text{aug}} = f(\mathbf{X}_c) + \mathbf{N}_c$$

where $\mathbf{F}_c^{\text{aug}}$ is the noise-augmented feature representation.

3. **Triplet Loss Integration:** The class-specific noise is optimized using the triplet loss. For each class $c$, we define an anchor $\mathbf{F}_c^{\text{anchor}}$ as the feature representation with noise, a positive sample $\mathbf{F}_c^{\text{positive}}$ from the same class, and a negative sample $\mathbf{F}_{c'}^{\text{negative}}$ from a different class $c'$. The triplet loss is defined as:

$$\mathcal{L}_{\text{triplet}} = \max\Big(0,\, d(\mathbf{F}_c^{\text{anchor}}, \mathbf{F}_c^{\text{positive}}) -d(\mathbf{F}_c^{\text{anchor}}, \mathbf{F}_{c'}^{\text{negative}}) + \alpha\Big)$$

where $d(\cdot)$ is a distance metric (e.g., Euclidean distance), and $\alpha$ is a margin that ensures the separation between classes is sufficiently large. The class-specific noise $\mathbf{N}_c$ modifies the anchor and positive samples by augmenting their feature representations. This modified triplet loss encourages the model to minimize the intra-class distance (i.e., between anchor and positive samples) while maximizing the inter-class distance (i.e., between anchor and negative samples).

The augmented triplet loss incorporating the class-specific noise can be written as:

$$\mathcal{L}_{\text{triplet}}^{\text{aug}} = \max\Big(0,\, d(f(\mathbf{X}_c) + \mathbf{N}_c, f(\mathbf{X}_c^{\text{positive}}) + \mathbf{N}_c) -d(f(\mathbf{X}_c) + \mathbf{N}_c, f(\mathbf{X}_{c'}^{\text{negative}}) + \mathbf{N}_{c'}) + \alpha\Big)$$

This loss function ensures that the noise $\mathbf{N}_c$ is optimized to highlight class-specific features, resulting in better inter-class separability.

4. **Fine-Tuning Pretrained Models with Class-Specific Noise:** The learned noise $\mathbf{N}_c$ is integrated into pretrained VLMs as a fine-tuning mechanism. The pretrained models can access the class-specific noise through the dictionary $\mathcal{D}_c$ either as an augmentation to the dataset or as a learnable parameter. Fine-tuning with the noise ensures that the pretrained models are optimized to focus on class-distinctive features, resulting in better classification performance under noise and masking conditions.

The feature representation of class $c$ after fine-tuning becomes:

$$\mathbf{F}_c^{\text{fine-tuned}} = f_{\text{VLM}}(\mathbf{X}_c) + \mathbf{N}_c$$

where $f_{\text{VLM}}$ represents the feature extraction function of the pretrained vision-language model (VLM), and $\mathbf{N}_c$ is the learned class-specific noise for class $c$.

5. **Evaluation and Hypothesis:** After learning the class-specific noise through triplet loss, the model is expected to achieve better performance compared to standard pretrained models. The learned noise helps preserve key class-defining features, reducing label dispersion and improving classification accuracy, especially in challenging settings where features are masked or perturbed.

**Triplet Loss:** In the context of our study, the Figure 1 illustrates the concept of triplet loss applied to our class-specific noise augmentation approach. Here, the Anchor represents a feature vector extracted from a frame belonging to a specific action class, for instance, *cricket shot*. The Positive denotes another frame from the same class, and the objective is to minimize the distance $d(\mathbf{F}_c^{\text{anchor}}, \mathbf{F}_c^{\text{positive}})$, thereby ensuring that features from the same class are closely aligned in the embedding space. Conversely, the Negative refers to a feature vector from a different action class, such as *tennis serve*. The goal is to maximize the distance $d(\mathbf{F}_c^{\text{anchor}}, \mathbf{F}_{c'}^{\text{negative}})$ between the anchor and the negative, effectively pushing apart features from different classes. The margin $\alpha$ acts as a threshold that enforces a minimum separation between the anchor and the negative, thereby enhancing the model's ability to differentiate between distinct action classes. In Figure 1 it can be seen Anchor (A) and Positive (P) are chosen as different frames of same action class (*Playing Cricket*) while a very similar class where model could get confused (*Tennis Swing*); as the hypothesis - model could end up learning non-class features from the same action class videos, for example a human body along with green grass or ground that are present in both the action classes.

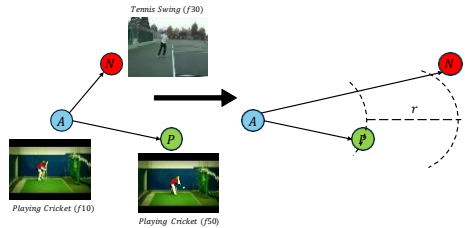

**Fig. 1.** Triplet Loss; The diagram including *A* for anchor, *N* for Negative class and *P* for positive class. The class images (video to frames) are taken from UCF101 dataset.

## 5    Implementation and Results

Following Section 4, we present the experiments in a task-wise format in this section. The experiments are conducted based on the specific tasks outlined, with results showcased for CLIP. To improve clarity and readability, we focus on a subset of action classes for more detailed ablation studies. These selected classes are chosen based on conditions that typically cause the models to struggle with label prediction, providing deeper insights into the model's limitations and performance.

### 5.1    Task 1

Observations from the experiments shows that for most of the actions, CLIP model is easily able to predict the ground truth label with some confidence, however it also predict other labels and sometimes with high confidence. For example, *Rope Climbing* action class predictions are shown by Figure 2. We release all other predictions for rest of the action classes in UCF101 in supplementary material (In Appendix).

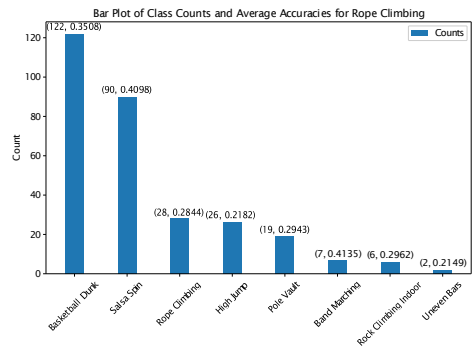

**Fig. 2.** The frequency histogram shows how *Rope Climbing* class is getting predicted with multiple other labels with high confidence, leading to *label dispersion* problem

### 5.2    Task 2

Three frequency histograms are plotted for each masking percentage (10, 30 and 50). The masking is done by randomly picking a pixel and then making it black. In short, calculate the number of pixels to be masked, and then mask those with black pixels. For example, for the action class *Apply Eye Makeup* predictions are shown in Figure 3. Observations reveal that as perturbation increases, the labels start to get diverse with more confidence, for example, with 50% masking, CLIP predicts the input image as *Shaving Beard* with more frequency and high confidence, while in a normal scenario, time limited human will easily classify it as *Apply Eye Makeup*. This observation also reveals, CLIP does not learn much in differentiating *Shaving Beard* is for male-specific features (mostly).

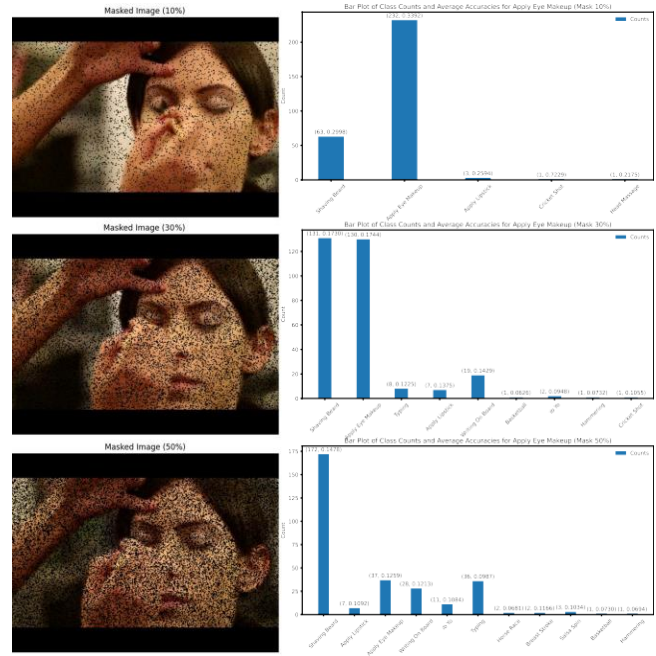

**Fig. 3.** Masked images for the ground truth class label *Apply Eye Makeup*. Images in left masked (10, 30 and 50 top to bottom) with corresponding predictions as frequency histogram in right

### 5.3   Tasks 3 and 4

Here, we perform feature-specific masking followed by predictions. Figures 4 and 5 showcases the results of this masking scenario. We conducted this experiment only for the *Cricket Shot* action class and left the other action classes for future testing, as they require hand-picking segmentation coordinates using SAM [17]. Observations reveal that none of the segmented images were labelled as the ground truth. Possible reasons for such result in (1), (2) and (5) where green grass, pitch and player are segmented respectively, from Figure 4 is because of the reason - that CLIP's vision model places more emphasis on background elements like grass and the overall scene context, rather than the specific action being performed. This leads to misclassifications as the model might associate these background features (such as the grass and pitch) with other action classes. Additionally, the segmentation of the player in (5) might not fully capture the motion or activity required for the correct labeling of *Cricket Shot*, as the action depends heavily on the player's dynamic pose, which could be lost during segmentation. Also, class indices 143, 14, 58, 72 and 86 seems like the most confused action classes where as indices 59 to 68 are very focused towards having the instrument (object) in the frame, due to which it was confident with the action class label.

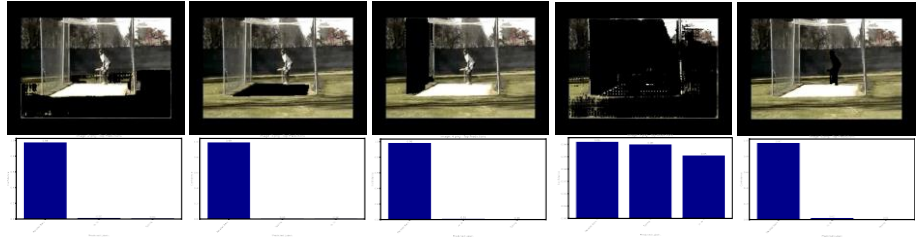

**Fig. 4.** Segmentation analysis for Task 3 on *Cricket Shot* - Chronologically (left to right) images are described as, (1) Green grass segmented, (2) Pitch segmented, (3) Cricket net segmented (4) Everything segmented from 2, 3 and 4, (5) Only player segmented.

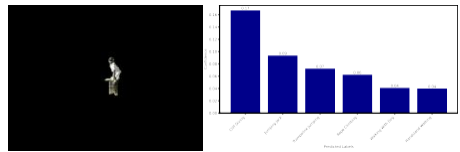

**Fig. 5.** Segmentation analysis for Task 4 on *Cricket Shot* - Segmenting everything but the player.

### 5.4   Task 5

Indeed class specific noise augmentation helps model to prevent label dispersion and with high confidence predictions. We have just proposed and explored this method for augmentation to the learning model, however, future scope of this also could be to present adversarial attacks and building robust models [13]. Also, more experiments can be carried using tasks 1 to 4 in order to verify methods for task 5. We present results in Figure 6.

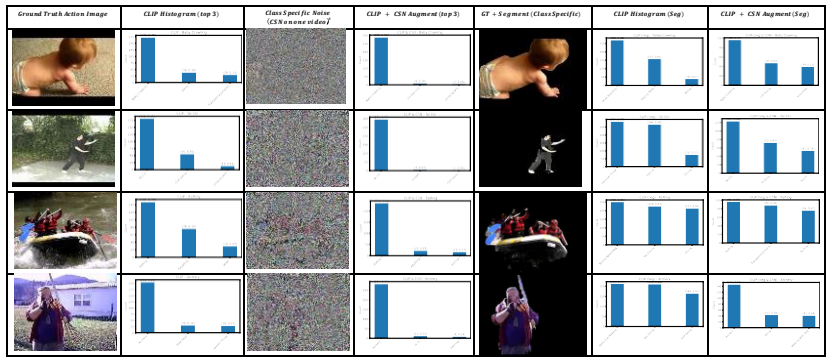

**Fig. 6.** Results show that class specific noise enhanced the predictions for both the cases; without SAM (left) and with SAM (rightost 3 columns)

## 6    Conclusion

The CLIP model's joint vision-language embedding space, pretrained generalisation, and ability to leverage textual prompts make it a versatile tool for human action recognition. This study explored the limitations and potential enhancements of the CLIP model for human action recognition by focusing on its behaviour under various masking strategies and introducing a novel class-specific noise augmentation approach. We proposed using class-specific noise, learned as a dictionary through a custom loss function, to mitigate these limitations and enhance the model's focus on class-defining features. Our experiments on the UCF101 dataset revealed that masking strategies such as random, feature-specific, and isolation masking significantly affect the CLIP model's performance. Specifically, the model often misclassified actions due to an inability to retain or prioritise the correct class-specific features, highlighting its sensitivity to disruptions in key visual patterns. Through frequency histogram analysis, we observe CLIP model exhibits inconsistent behaviour when exposed to different masking strategies, including (a) random percentage-based and shape-based black masking at 10%, 30%, and 50%, (b) feature-specific masking to remove bias-inducing features, and (c) isolation masking, where class-specific features are selectively preserved and labelled. This augmentation strategy improved classification confidence and accuracy, addressing the label dispersion problem and demonstrating the utility of incorporating targeted noise in action recognition tasks. Our finding shows that CLIP has the potential for a robust framework for applications in healthcare, including real-time monitoring and decision-making in critical medical scenarios.

## 7    Challenges and Future Research

While CLIP excels in generalisation, its performance may degrade in specific domains (e.g., healthcare) due to domain-specific biases. Techniques such as class-specific noise augmentation, task-specific fine-tuning, and temporal modelling can be employed to enhance its reliability for action recognition in healthcare. Future studies should evaluate the effectiveness of class-specific noise on diverse human action recognition datasets, such as Kinetics-400 or HMDB51, to assess its generalizability and robustness. We will also explore this research on healthcare and medical human action recognition datasets. Additionally, combining visual and audio modalities could further improve recognition accuracy. By addressing these directions, the class-specific noise approach has the potential to significantly advance the state of human action recognition, contributing to the broader goal of building more robust and interpretable AI systems.

**Acknowledgements**  The research presented in this paper has benefited from the Experimental Infrastructure for Exploration of Exascale Computing (eX3), which is financially supported by the Research Council of Norway under contract 270053.

# Appendix

In this section we outline all class wise results for task 1 in Tables 1, 2 and 3. Moreover we also release handpicked interesting results in Table 4.

| Index | Class Label | Preds. |
|---|---|---|
| 1 | Apply Eye Makeup | 1 (300, 0.96) |
| 2 | Apply Lipstick | 2 (162, 0.44), 20 (116, 0.51), 1 (22, 0.47) |
| 3 | Archery | 3 (291, 0.57), 31 (5, 0.27), 56 (2, 0.23), 36 (2, 0.27) |
| 4 | Baby Crawling | 4 (300,0.96) |
| 5 | Balance Beam | 69 (153, 0.45), 96 (147, 0.49) |
| 6 | Band Marching | 6 (300, 0.93) |
| 7 | Baseball Pitch | 7 (300, 0.39) |
| 8 | Basketball | 48 (125, 0.27), 94 (87, 0.32), 51 (56, 0.22), 29 (19, 0.22), 68 (6, 0.26), 85 (5, 0.22), 79 (2, 0.15) |
| 9 | Basketball Dunk | 9 (300,0.81) |
| 10 | Bench Press | 10 (300,0.85) |
| 11 | Biking | 11 (300, 0.92) |
| 12 | Billiards | 12 (89, 0.54), 90 (211,0.61) |
| 13 | Blow Dry Hair | 13 (76, 0.59), 46 (78, 0.59), 39 (67, 0.40), 41 (35, 0.36), 67 (15, 0.29), 56 (14, 0.37), 26 (8, 0.21), 32 (2, 0.19), 52 (2, 0.14), 33 (1, 0.13), 79 (1, 0.12), 81 (1, 0.44) |
| 14 | Blowing Candles | 14 (281, 0.96), 9 (5,0.07), 77 (4, 0.08), 46 (3, 0.15), 84 (2, 0.17), 57 (2, 0.08), 45 (1, 0.13), 24 (1, 0.08), 94 (1, 0.23) |
| 15 | Body Weight Squats | 15 (191, 0.54), 52 (109, 0.52) |
| 16 | Bowling | 16 (274, 0.81), 30 (11, 0.37), 90 (8, 0.26), 51 (3, 0.31), 46 (2, 0.36), 57 (2, 0.33) |
| 17 | Boxing Punching Bag | 17 (162, 0.53), 18 (123, 0.54), 9 (15, 0.35) |
| 18 | Boxing Speed Bag | 18 (105, 0.47), 21 (18, 0.29), 17 (149, 0.45), 25 (24, 0.39), 99 (4, 0.28) |
| 19 | Breast Stroke | 19 (36, 0.40), 32 (230, 0.56), 26 (34, 0.32) |
| 20 | Brushing Teeth | 20 (19, 0.45), 78 (281, 0.61) |
| 21 | Clean And Jerk | 21 (300, 0.92) |
| 22 | Cliff Diving | 22 (192, 0.80), 38 (46, 0.39), 94 (20, 0.31), 51 (15, 0.31), 45 (9, 0.21), 68 (7, 0.19), 57 (5, 0.12), 24 (3, 0.11), 79 (3, 0.21) |
| 23 | Cricket Bowling | 23 (96, 0.56), 24 (204, 0.59) |
| 24 | Cricket Shot | 45 (300, 0.48) |
| 25 | Cutting In Kitchen | 25 (300, 0.98) |
| 26 | Diving | 26 (300, 0.80) |
| 27 | Drumming | 27 (300, 0.89) |
| 28 | Fencing | 28 (235,0.44), 77 (35, 0.19), 30 (7, 0.14), 91 (7, 0.19), 84 (6, 0.31), 90 (5, 0.16), 48 (5, 0.11) |
| 29 | Field Hockey Penalty | 29 (300, 0.98) |
| 30 | Floor Gymnastics | 30 (242, 0.54), 57 (29, 0.43), 51 (27, 0.44), 90 (2, 0.38) |

**Table 1.** Class labels and predictions (1-30) Task 1

| Index | Class Label | Preds. |
|---|---|---|
| 31 | Frisbee Catch | 31 (132, 0.40), 29 (145, 0.50), 36 (12, 0.20), 45 (4, 0.16), 93 (4, 0.22), 84 (3, 0.31) |
| 32 | Front Crawl | 32 (300, 0.97) |
| 33 | Golf Swing | 33 (36, 0.59), 84 (205, 0.58), 29 (31, 0.47), 51 (18, 0.34), 45 (10, 0.32) |
| 34 | Haircut | 39 (122, 0.45), 13 (118, 0.46), 86 (22, 0.20), 25 (14, 0.21), 50 (10, 0.20), 56 (4, 0.15), 67 (4, 0.08), 2 (2, 0.13), 62 (2, 0.14), 78 (2, 0.21) |
| 35 | Hammering | 75 (300, 0.95) |
| 36 | Hammer Throw | 36 (175, 0.41), 43 (5, 0.28), 68 (97, 0.30), 45 (15, 0.26), 69 (5, 0.20), 40 (3, 0.18) |
| 37 | Handstand Pushups | 37 |
| 38 | Handstand Walking | 38 (296, 0.58), 30 (4, 0.35) |
| 39 | Head Massage | 39 (300, 0.89) |
| 40 | High Jump | 40 (57, 0.44), 51 (162, 0.44), 68 (70, 0.54), 45 (11, 0.29) |
| 41 | Horse Race | 41 (300, 1.00) |
| 42 | Horse Riding | 42 (267, 0.59), 40 (22, 0.28), 41 (7, 0.29), 48 (4, 0.21) |
| 43 | Hula Hoop | 46 (175, 0.38), 77 (125, 0.35) |
| 44 | Ice Dancing | 44 (300, 0.99) |
| 45 | Javelin Throw | 45 (13, 0.64), 51 (287, 0.57) |
| 46 | Juggling Balls | 46 (140, 0.48), 79 (76, 0.32), 91 (33, 0.34), 67 (28, 0.27), 84 (11, 0.35), 48 (6, 0.19), 59 (4, 0.16), 88 (1, 1, 0.14), 51 (1, 0.14) |
| 47 | Jumping Jack | 47 (4, 0.12), 46 (94, 0.25), 38 (78, 0.20), 84 (66, 0.16), 94 (29, 0.22), 91 (14, 0.23), 48 (8, 0.16), 9 (4, 0.52), 47 (4, 0.12), 58 (3, 0.25) |
| 48 | Jump Rope | 48 (260, 0.57), 84 (28, 0.40), 77 (9, 0.33), 91 (3, 0.33) |
| 49 | Kayaking | 49 (266, 0.85), 88 (30, 0.55), 46 (2, 0.24), 74 (2, 0.53) |
| 50 | Knitting | 50 (56, 0.54), 25 (236, 0.55), 86 (4, 0.22), 75 (4, 0.20) |
| 51 | Long Jump | 51 (184, 0.67), 68 (116, 0.61) |
| 52 | Lunges | 52 (257, 0.59), 48 (43, 0.44) |
| 53 | Military Parade | 53 (300, 0.91) |
| 54 | Mixing | 25 (291, 0.21), 11 (6, 0.15), 27 (3, 0.14) |
| 55 | Mopping Floor | 55 (300, 0.99) |
| 56 | Nunchucks | 56 (103, 0.45), 91 (171, 0.37), 52 (14, 0.23), 72 (4,0.27), 48 (4, 0.27), 77 (3, 0.26), 60 (1, 0.22) |
| 57 | Parallel Bars | 57 (103, 0.63), 68 (197, 0.63) |
| 58 | Pizza Tossing | 58 (205, 0.43), 46 (64, 0.43), 30 (9, 0.24), 9 (7, 0.38), 99 (5, 0.18), 72 (5, 0.16), 25 (3, 0.11), 17 (2, 0.21) |
| 59 | Playing Cello | 59 (300,0.98) |
| 60 | Playing Daf | 46 (188, 0.67), 61 (112, 0.71) |
| 61 | Playing Dhol | 61 (300, 0.94) |
| 62 | Playing Flute | 62 (300, 0.95) |
| 63 | Playing Guitar | 63 (300, 0.95) |
| 64 | Playing Piano | 64 (300, 0.97) |
| 65 | Playing Sitar | 65 (300, 0.94) |
| 66 | Playing Tabla | 66 (300, 0.95) |
| 67 | Playing Violin | 67 (300, 0.91) |
| 68 | Pole Vault | 68 (296, 0.92), 94 (4, 0.72) |
| 69 | Pommel Horse | 69 (236, 0.47), 96 (27, 0.34), 30 (20, 0.29), 40 (5, 0.22), 45 (4, 0.29), 68 (3, 0.21), 51 (3, 0.25), 90 (2, 0.29) |

**Table 2.** Class labels and predictions (31-69) Task 1

| Index | Class Label | Preds. |
|---|---|---|
| 70 | Pull Ups | 70 (9, 0.52), 57 (291, 0.69) |
| 71 | Punch | 71 (165, 0.55), 17 (134, 0.50), 18 (1, 0.33) |
| 72 | Push Ups | 72 (223, 0.52), 94 (20, 0.41), 87 (20, 0.52), 82 (17, 0.56), 51 (17, 0.40), 57 (3, 0.44) |
| 73 | Rafting | 73 (300, 0.99) |
| 74 | Rock Climbing Indoor | 74 (300, 0.99) |
| 75 | Rope Climbing | 75 (122, 0.35), 77 (90, 0.41), 75 (28, 0.28), 40 (26, 0.22), 68 (19, 0.29), 6 (7, 0.41), 74 (6, 0.30), 96 (2, 0.21) |
| 76 | Rowing | 76 (286, 0.91), 73 (7, 0.50), 27 (5, 0.25), 58 (2, 0.17) |
| 77 | Salsa Spin | 77 (300, 0.86) |
| 78 | Shaving Beard | 78 (268, 0.66), 70 (9, 0.25), 27 (9, 0.22), 58 (7, 0.28), 10 (7, 0.30) |
| 79 | Shotput | 79 (300, 0.79) |
| 80 | Skate Boarding | 80 (65, 0.48), 51 (58, 0.11), 11 (34, 0.24), 91 (12, 0.11), 77 (4, 0.12) |
| 81 | Skiing | 81 (152, 0.52), 51 (115, 0.49), 68 (23, 0.48), 82 (10, 0.51) |
| 82 | Ski Jet | 82 (5, 0.75), 88 (295, 0.88) |
| 83 | Sky Diving | 83 (300, 0.75) |
| 84 | Soccer Juggling | 84 (229, 0.38), 46 (40, 0.34), 48 (23, 0.26), 85 (8, 0.33) |
| 85 | Soccer Penalty | 85 (300, 0.89) |
| 86 | Still Rings | 68 (104, 0.45), 75 (102, 0.51), 96 (50, 0.33), 57 (20, 0.27), 40 (8, 0.25), 69 (5, 0.28), 9 (3, 0.23), 30 (2, 0.18), 48 (2, 0.18), 21 (2, 0.24), 46 (2,0.18) |
| 87 | Sumo Wrestling | 87 (300, 1.00) |
| 88 | Surfing | 88 (300, 0.97) |
| 89 | Swing | 89 (19, 0.26), 94 (230, 0.35), 48 (40, 0.21), 31 (3, 0.13), 4 (8, 0.23) |
| 90 | Table Tennis Shot | 90 (300, 1.00) |
| 91 | Tai Chi | 91 (204, 0.65), 48 (78, 0.43), 98 (18, 0.30) |
| 92 | Tennis Swing | 92 (298, 0.84), 36 (2, 0.20) |
| 93 | Throw Discus | 93 (30,0.29), 51 (135, 0.38), 45 (113, 0.30), 23 (15, 0.33), 9 (7,0.32) |
| 94 | Trampoline Jumping | 94 (300, 1.00) |
| 95 | Typing | 95 (300, 0.59) |
| 96 | Uneven Bars | 96 (286, 0.58), 68 (10, 0.35), 69 (3, 0.43), 36 (1, 0.15) |
| 97 | Volleyball Spiking | 97 (121, 0.38), 8 (88, 0.45), 9 (76, 0.56), 84 (6, 0.20), 31 (3, 0.17), 28 (3, 0.18), 38 (3, 0.15) |
| 98 | Walking With Dog | 98 (283, 0.66), 77 (8, 0.35), 46 (5, 0.20), 91 (3, 0.34), 94 (1, 0.20) |
| 99 | Wall Pushups | 99 (254, 0.27), 55 (29, 0.18), 43 (10, 0.15), 91 (7, 0.16) |
| 100 | Writing On Board | 100 (300, 0.64) |
| 101 | Yo Yo | 91 (206, 0.53), 46 (34, 0.34), 67 (23, 0.43), 90 (13, 0.37), 48 (13, 0.28), 77 (8, 0.26), 98 (2, 0.28), 87 (1, 0.26) |

**Table 3.** Class labels and predictions (69-101) Task 1

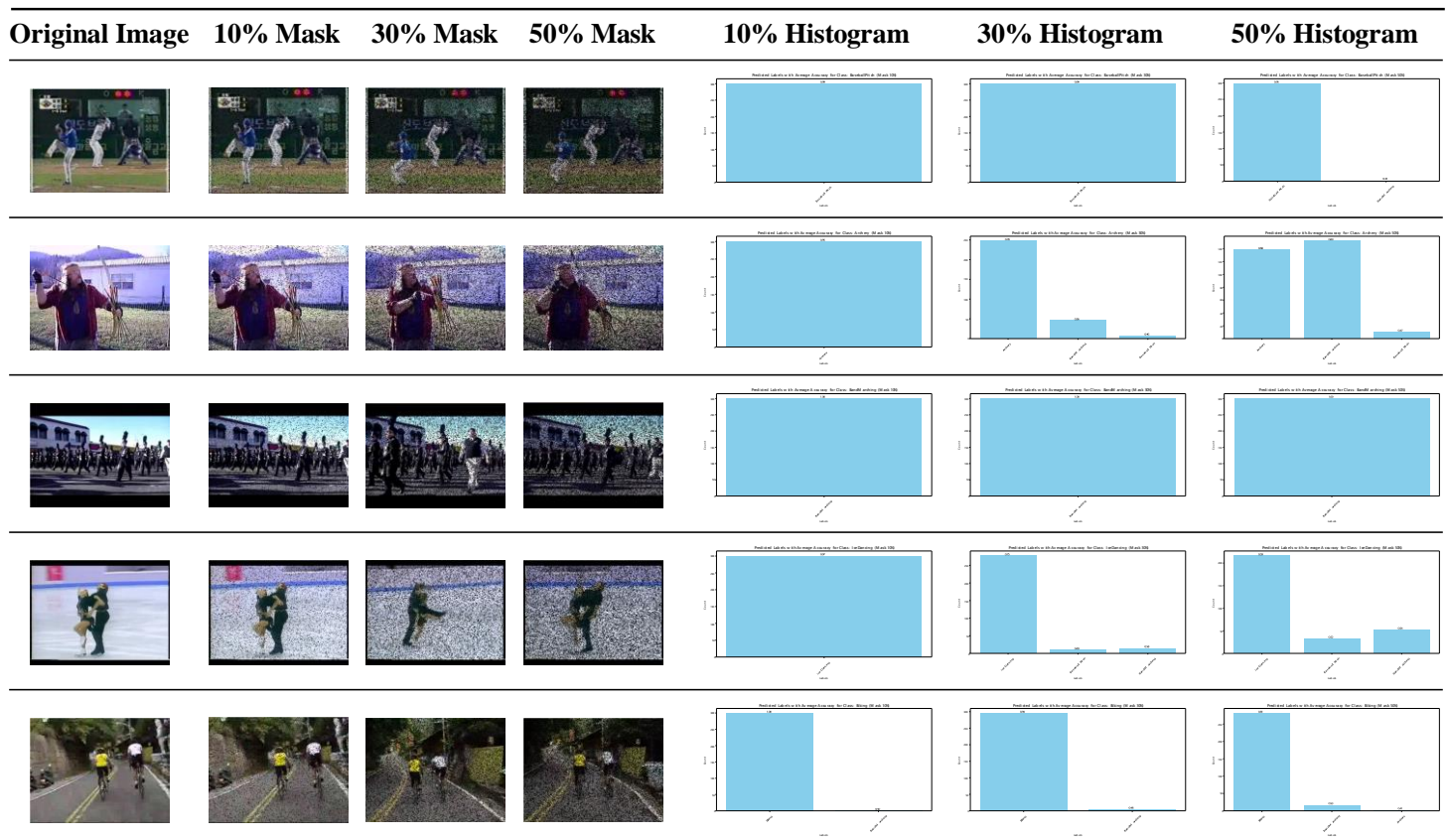

| Original Image | 10% Mask | 30% Mask | 50% Mask | 10% Histogram | 30% Histogram | 50% Histogram |
| --- | --- | --- | --- | --- | --- | --- |

**Table 4.** Handpicked Comparison of Original and Masked Images with Corresponding Histograms for Task 2; Observations reveal for *Baseball Pitch* CLIP was actually giving best performance even with the 50% perturbation, *Archery* it starts to get confused with 30% perturbation. For rest of the actions also it gets confused easily with 50% perturbation. Here witth his particular experiment we kept only four labels in the ground truth labels for CLIP to take as query. But above all, time limited humans with enough information would label it correctly